%% file: paper.tex
\documentclass[twoside,leqno,twocolumn]{article}
\usepackage[letterpaper]{geometry}

\usepackage{ltexpprt}
\usepackage{hyperref}

\newif\ifapx
\apxtrue 

\newcommand{\ourmaintitle}{Data is Moody: Discovering Data Modification Rules from Process Event Logs}
\newcommand{\ourtitle}{\ourmaintitle}
\newcommand{\ourmethod}{\textsc{Moody}\xspace}
\newcommand{\oururl}{\url{https://eda.rg.cispa.io/prj/moody/}}
\newcommand{\codeurl}{\oururl}

\makeatletter
\newif\if@restonecol
\makeatother

\usepackage[ruled, vlined, nofillcomment, linesnumbered]{algorithm2e}
\usepackage{cite}
\usepackage{amsmath,amssymb,amsfonts}
\usepackage{graphicx}
\usepackage{textcomp}
\usepackage{xcolor}
\def\BibTeX{{\rm B\kern-.05em{\sc i\kern-.025em b}\kern-.08em
    T\kern-.1667em\lower.7ex\hbox{E}\kern-.125emX}}
\usepackage[notheorems]{eda}  
\pgfplotsset{compat=1.14}
\usepackage{balance}

\graphicspath{ {./figs/} }

\ifpdf
\usetikzlibrary{calc}
\usetikzlibrary{external}
\fi

\usepackage{todonotes} 
\usepackage{letltxmacro}
\LetLtxMacro{\oldtodo}{\todo}
\renewcommand{\todo}[2][]{\tikzexternaldisable\oldtodo[#1]{#2}\tikzexternalenable}
\LetLtxMacro{\oldmissingfigure}{\missingfigure}
\renewcommand{\missingfigure}[2][]{\tikzexternaldisable\oldmissingfigure[{#1}]{#2}\tikzexternalenable}

\usepackage[notheorems]{eda}  

\input{defines}

\begin{document}
	\newcommand\relatedversion{}

	\title{\Large\ourtitle\relatedversion}

	\author{
		Marco Bjarne Schuster\thanks{\rule{0pt}{1.1em}Airbus Operations GmbH, Bremen, Germany.\newline
		\-\hspace{16pt}\texttt{marco.schuster@airbus.com}}
		\and
		Boris Wiegand\thanks{\rule{0pt}{1.1em}Stahl-Holding-Saar, Dillingen, Germany.\newline
		\-\hspace{16pt}\texttt{boris.wiegand@stahl-holding-saar.de}}
		\and
		Jilles Vreeken\thanks{\rule{0pt}{1.1em}CISPA Helmholtz Center for Information Security, Germany.\newline
		\-\hspace{16pt}\texttt{jv@cispa.de}}
	}

	\date{}

	\maketitle


	\begin{abstract} \small\baselineskip=9pt
	\input{abstract}

	\end{abstract}

	\input{introduction}
	\input{related}
	\input{preliminaries}
	\input{theory}
	\input{algorithm}
	\input{experiments}
	\input{discussion}

	\input{conclusion}


	\bibliographystyle{abbrv}
	\bibliography{bib/abbrev,bib/bib-jilles,bib/bib-paper}

	\ifapx
	\appendix
	\input{appendix}
	\fi

\end{document}

%% file: defines.tex
\SetKwComment{tcpas}{\{}{\}}
\SetCommentSty{textnormal}
\SetArgSty{textnormal}
\SetKwRepeat{Do}{do}{while}
\SetKw{False}{false}
\SetKw{True}{true}
\SetKw{Null}{null}
\SetKwInOut{Output}{output}
\SetKwInOut{Input}{input}
\SetKw{AND}{and}
\SetKw{OR}{or}
\SetKw{Continue}{continue}

\pgfplotsset{
	discard if not/.style 2 args={
		x filter/.code={
			\edef\tempa{\thisrow{#1}}
			\edef\tempb{#2}
			\ifx\tempa\tempb
			\else
			
			\fi
		}
	}
}

\usetikzlibrary{arrows.meta}
\usetikzlibrary{shapes.multipart}
\usetikzlibrary{positioning}

\usepackage{pifont}
\usepackage[linewidth=1pt]{mdframed}

\definecolor{backOrange}{HTML}{FFE6CC}
\definecolor{backBlue}{HTML}{DAE8FC}
\definecolor{backPurple}{HTML}{E1D5E7}
\definecolor{backGreen}{HTML}{D5E8D4}
\definecolor{backTurquoise}{HTML}{B0E3E6}

\preto\tabular{\setcounter{rulenumbers}{0}}
\newcounter{rulenumbers}
\preto\figure{\setcounter{eventnumbers}{0}}
\newcounter{eventnumbers}

\newcommand{\refline}[1]{ln.\ \ref*{#1}}
\newcommand{\refrule}[1]{R\ref*{#1}}
\newcommand{\refevent}[1]{E\ref*{#1}}

\newcommand{\LN}{L_\mathbb{N}}
\newcommand{\LR}{\ensuremath{L_\mathbb{R}}}
\newcommand{\paramNc}{N_c}
\newcommand{\paramNu}{N_u}
\newcommand\rulenumber{\refstepcounter{rulenumbers}R\arabic{rulenumbers}:}
\newcommand{\rulenumbersautorefname}{R\kern-4pt}
\newcommand\eventnumber{\refstepcounter{eventnumbers}E\arabic{eventnumbers}}
\newcommand{\eventnumbersautorefname}{E\kern-4pt}
\newcommand{\mycaption}[2]{\caption[#1]{[\textbf{#1}] #2}}
\newcommand{\checkmarkCode}[0]{\ding{51}}
\newcommand{\crossCode}[0]{\ding{55}}

\newcommand{\nrOfTraces}{\ensuremath{\abs{D}}}
\newcommand{\nrOfEvents}{\ensuremath{||D||}}
\newcommand{\setOfCategoricalVariables}{\ensuremath{V_{\text{cat}}}}
\newcommand{\setOfNumericalVariables}{\ensuremath{V_{\text{num}}}}

\DeclareMathOperator{\dom}{dom}

\DeclareMathOperator*{\argmax}{arg\, max}
\DeclareMathOperator*{\argmin}{arg\, min}
\DeclareMathOperator{\freq}{fr}
\DeclareMathOperator*{\usg}{usg}
\DeclareMathOperator*{\supp}{supp}

\newcommand{\fonescore}{\ensuremath{\text{F}_1~\text{score}}\xspace}
\newcommand{\fonescores}{\ensuremath{\text{F}_1~\text{scores}}\xspace}

\newcommand{\ssdpp}{\textsc{SSD\texttt{++}}\xspace}
\newcommand{\turs}{\textsc{Turs}\xspace}
\newcommand{\classy}{\textsc{Classy}\xspace}

\newcommand{\sepsis}{Sepsis\xspace}
\newcommand{\trafficfines}{Traffic Fines\xspace}
\newcommand*{\doi}[1]{\href{http://dx.doi.org/#1}{doi: #1}}

%% file: abstract.tex
Although event logs are a powerful source to gain insight about the behavior of the underlying business process,
existing work primarily focuses on finding patterns in the activity sequences of an event log,
while ignoring event attribute data.
Event attribute data has mostly been used to predict
event occurrences and process outcome,
but the state of the art neglects to mine succinct and interpretable rules
how event attribute data changes during process execution.
Subgroup discovery and rule-based classification approaches lack the ability to capture the
sequential dependencies present in event logs,
and thus lead to unsatisfactory results with limited insight into the process behavior.

Given an event log, we are interested in finding accurate yet succinct and interpretable if-then rules
how the process modifies data.
We formalize the problem in terms of the Minimum Description Length (MDL) principle,
by which we choose the model with the best lossless description of the data.
Additionally, we propose the greedy \ourmethod algorithm to efficiently search for rules.
By extensive experiments on both synthetic and real-world data,
we show \ourmethod indeed finds compact and interpretable rules,
needs little data for accurate discovery, and is robust to noise.

%% file: introduction.tex
\section{Introduction}
\label{sec:intro}

Given a process event log, process mining \cite{van_der_aalst_process_2016}
provides a better understanding of the underlying process and enables downstream tasks such as monitoring, anomaly detection, simulation, and optimization.
Existing work, however, focuses on discovering patterns of event activities,
but neglects how event attribute data changes during the process.
For instance, a textile company may have an ordering process with the activities
\emph{Request}, \emph{Place}, \emph{Delay} and \emph{Receive}.
Process discovery algorithms \cite{augusto:2017:splitminer,sommers:2021:pd_gnn} only infer a graph of event activities,
where nodes refer to activities and edges visualize the flow of execution.
However, how data changes over time is crucial for understanding the process as we show in Figure~\ref{fig:introduction-example}.
In our textile example, we only know the price and the delivery date of an order after placing the order.
Delaying the delivery of an order, e.g.\ due to a shortness of supplies, changes the delivery date.

Surprisingly, data has only been used to predict the occurrences of event activities or the outcome of processes \cite{taymouri:2021:deep,wiegand:2022:consequence}.
To the best of our knowledge, none of the existing work mines interpretable rules for data modifications,
and related work does not satisfactorily transfer to our problem.
Subgroup discovery \cite{proenca:2022:ssdpp} and rule-based prediction methods \cite{yang:2022:turs}
lack the ability to model the sequential dependencies present in event logs,
and thus lead to unsatisfactory results with limited insight into the process behavior.

Given an event log, we are interested in finding accurate
yet succinct and interpretable if-then rules how the process modifies data.
To this end, we formalize the problem in terms of the Minimum Description Length (MDL) principle,
by which we choose the model with the best lossless description of the data.
Additionally, we propose our method \ourmethod, which is short for \textsc{Mo}dification rule \textsc{D}iscover\textsc{y}, to efficiently search for rule models in practice.
Starting with an empty set, we greedily add the best compressing rule to the model,
until we no longer find a rule that improves our MDL score.
Through extensive experiments on both synthetic and real-world data,
we show \ourmethod indeed finds succinct and interpretable rules,
needs little data for accurate discovery, and is robust to noise.

\begin{figure}
	\centering
	\tikzsetnextfilename{introduction-example}
	\begin{tikzpicture}[node distance=2cm,bend angle=45,auto]
		\usetikzlibrary{arrows.meta}
		\node[draw,rectangle] (req) {Request};
		\node[draw,rectangle] (place) [right of=req] {Place};
		\node[draw,rectangle] (del) [right of=place] {Delay};
		\node[draw,rectangle] (rec) [right of=del] {Recieve};
		\path[-{Latex[length=2mm,width=2mm]}] (req) edge (place);
		\path[-{Latex[length=2mm,width=2mm]}] (place) edge (del);
		\path[-{Latex[length=2mm,width=2mm]}] (del) edge (rec);
		\path[-{Latex[length=2mm,width=2mm]}] (place) edge[bend left] (rec);
		\path[-{Latex[length=2mm,width=2mm]}] (del) edge[out=120, in=60, distance=0.8cm] (del);
		\node[draw,rectangle,fill=backBlue] (reqOrder) [below of=req, node distance=1.2cm, text width=1.5cm] {\textbf{product:}\\t-shirt\\\textbf{amount:}\\100};
		\node[draw,rectangle,fill=backBlue] (placeOrder) [below of=place, node distance=2cm, text width=1.5cm] {\textbf{product:}\\t-shirt\\\textbf{amount:}\\100\\\textbf{price:}\\\textdollar 1000\\\textbf{delivery:}\\May 1$^{\text{st}}$};
		\node[draw,rectangle,fill=backBlue] (delOrder) [below of=del, node distance=2cm, text width=1.5cm] {\textbf{product:}\\t-shirt\\\textbf{amount:}\\100\\\textbf{price:}\\\textdollar 1000\\\textbf{delivery:}\\May 9$^{\text{th}}$};
		\node[draw,rectangle,fill=backBlue] (recOrder) [below of=rec, node distance=2cm, text width=1.5cm] {\textbf{product:}\\t-shirt\\\textbf{amount:}\\100\\\textbf{price:}\\\textdollar 1000\\\textbf{delivery:}\\May 9$^{\text{th}}$};
	\end{tikzpicture}
	\mycaption{Processes change data}{
		Exemplary process for ordering textiles with activities \emph{Request}, \emph{Place}, \emph{Delay} and \emph{Receive}.
		Arrows indicate the flow of events.
		Further, we show how the data of an exemplary order changes throughout the process.
	}
	\label{fig:introduction-example}
\end{figure}

The contributions of our paper are as follows. We
\begin{itemize}[noitemsep,topsep=0pt]
	\item[(a)] formulate the problem of finding data modification rules from event logs,
	\item[(b)] formalize the problem using the Minimum Description Length (MDL) principle,
	\item[(c)] propose the \ourmethod algorithm to efficiently find accurate yet succinct data modification rules,
	\item[(d)] run extensive experiments on both synthetic and real-world data,
	\item[(e)] make code, data and appendix publicly available.\!\footnote{\codeurl}
\end{itemize}
Our paper is structured as usual. 

%% file: related.tex
\section{Related Work}\label{sec:related}

While the problem of finding interpretable data modification rules from process event logs
has been neglected so far, related work on similar problems exists.
Krismayer\cite{krismayer:2020:automatic} discovers if-then rules for data modification from software execution logs.
However, he only creates a large set of potentially redundant candidates,
which must be manually filtered by domain experts.
Other work \cite{walkinshaw:2016:efsm, foster:2021:efsm} infers extended finite state machines
from software execution logs.
Since business process event logs in contrast to software event logs are usually noisy and contain nondeterministic behavior,
these methods are not applicable to our problem.

While process mining \cite{van_der_aalst_process_2016} focuses on business process event logs,
most of the work only models the flow of process activities,
and little work deals with additional data.
Mannhardt et al. \cite{mannhardt:2016:conformance} and Mozafari et al. \cite{mozafari:2021:privacy} detect at which event data changes,
but do not model conditions or update values for these changes.
Sch{\"o}nig et al. \cite{schonig:2016:discovery} find rules which cover data modifications.
However, since they rely on support and confidence to filter a large set of candidate rules,
their method suffers from pattern explosion, i.e., it finds many redundant rules.

Rule-based prediction is closely related to our problem.
\classy \cite{proencca:2020:classy} and its successor \turs \cite{yang:2022:turs} find classification rules by minimizing an MDL score.
Both methods, however, require defining features and predicted variable beforehand,
whereas we are interested in finding relationships without any initial knowledge of the data.
Similarly, subgroup discovery algorithms such as \ssdpp\cite{proenca:2022:ssdpp} find rules for differently behaving subgroups of a given dataset.
None of these methods are able to model sequential relationships present in event logs.

In contrast to the above, \ourmethod finds compact and interpretable rules
for data modifications from process event logs,
needs little data for accurate discovery, and is robust to noise.

%% file: preliminaries.tex
\section{Preliminaries}
\label{sec:prelim}

Before formalizing the problem, we introduce preliminary
concepts and notation we use in the paper.

\subsection{Notation for Data Modification Rules}
As input for finding data modification rules, we consider an event log or dataset $D$ collecting traces of a single process.
Each trace refers to an instance of the process, such as a specific customer order, and consists of an event sequence.
We describe data attributes of events by a set of numerical and categorical variables $V$.

To model how a process modifies these variables, we use different types of update rules.
For a categorical variable $v \in V$, we write $v \in \{\alpha, \beta, \dots\}$,
i.e., $v$ takes one of the values in the set.
For a numerical variable $v \in V$, we can set $v$ to a specific value, $v = \alpha$, or to a range of values, $v \in [\alpha, \beta]$.
We further denote relative changes by $v = v + \alpha$, $v = v + [\alpha, \beta]$, and $v = \alpha \cdot v$.

Updates typically only occur in certain circumstances.
For instance, the price of an order may be dependent on the order volume,
where a higher volume gives discount.
Therefore, we model conditions for update rules.
In the simplest case, we check for a specific value $v = \alpha$ or $v \neq \alpha$.
Further, we test lower and upper bounds of numerical values by $v \le \alpha$ and $v \ge \alpha$.
Finally, we can condition on value transitions between the last and the current event with $v: \alpha \rightarrow \beta$.

We combine a condition $c$ and an update rule $u$ into a data modification rule \textbf{IF} $c$ \textbf{THEN} $u$.
To join multiple rules into a model $M$ that covers the full complexity of the process,
we use an unordered set of rules,
 since this allows for an independent interpretation of each rule \cite{yang:2022:turs}.
Furthermore, since we want to avoid contradictory predictions of our model, we only allow acyclic models.
For instance, if we condition on $v_1$ to update $v_2$, we are not allowed to do the reverse in the same model.

\subsection{Minimum Description Length}

We use the Minimum Description Length (MDL) principle \cite{rissanen:78:mdl,grunwald:07:book} for model selection.
MDL defines the best model as the one with the shortest lossless description of the given data.
Formally, the best model minimizes $L(M) + L(D \mid M)$,
in which $L(M)$ is the length in bits of the description of $M$, and $L(D \mid M)$ is the length of the data encoded with the model.
This form of MDL is known as two-part or crude MDL.
Although one-part or refined MDL has stronger theoretical guarantees,
it is only computable in specific cases \cite{grunwald:07:book}.
Therefore, we use two-part MDL.
In MDL, we only compute code lengths, but are not concerned with actual code words.
Next, we formalize our problem in terms of MDL.

%% file: theory.tex
\section{MDL for data modifications}
\label{sec:theory}

From an event log $D$, we aim to find a model $M$ of data modification rules,
which accurately describe the data, yet are as succinct as possible,
such that domain experts gain insight into the process.
Since real-world data is usually noisy, we need a robust model selection criterion.
Therefore, we formalize the problem in terms of the MDL principle.
To this end, we define length of the data encoding $L(D \mid M)$,
length of the model encoding $L(M)$, and finally give a formal problem definition.

\begin{figure}
	\tikzsetnextfilename{data-encoding-example}
	\begin{mdframed}
		\centering
		\begin{minipage}[t]{0.62\textwidth}
			\centering
			\textbf{Model:}\\
			\vspace{0.4em}
			\setlength\tabcolsep{2pt} 
			\begin{tabular}{rl}
				\rulenumber \label{rule:cat-prediction-1} & \textbf{IF} \textit{amount} $=$ 10 \\
				& \textbf{THEN} \textit{vendor} $=$ C \\
				\rulenumber \label{rule:cat-prediction-2} & \textbf{IF} \textit{product} $=$ bag \\
				& \textbf{THEN} \textit{vendor} $\in$ \{B, C\} \\
			\end{tabular}
		\end{minipage}%
		\begin{minipage}[t]{0.4\textwidth}
			\centering
			\textbf{Code streams:}\\
			\raggedright
			\vspace{0.4em}
			$C_m:$ \checkmarkCode$_{\refrule{rule:cat-prediction-2}}$, \checkmarkCode$_{\refrule{rule:cat-prediction-1}}$, \crossCode$_{\refrule{rule:cat-prediction-1}}$\\[0.4em]
			$C_v:$ \colorbox{backOrange}{C}, \colorbox{backBlue}{A}, \colorbox{backPurple}{A}\\[0.4em]
			$C_r:$ \refrule{rule:cat-prediction-1}\\
		\end{minipage}\\
		\vspace{0.5em}
		\textbf{Trace:}\\
		\tikzexternalexportnextfalse
		\begin{tikzpicture}
			\usetikzlibrary{shapes}
			\usetikzlibrary{arrows.meta}
			\usetikzlibrary{positioning}
			\node[rectangle split, rectangle split parts=3, rectangle split part align={center,center},
			draw, rectangle split part fill={white!0,white!0,backOrange}] (first)
			{
				bag
				\nodepart{two}
				20
				\nodepart{three}
				C
			};
			\node[rectangle split, rectangle split parts=3, rectangle split part align={center,center},
			draw, rectangle split part fill={white!0,white!0,white!0}, right=0.5 cm of first] (second)
			{
				bag
				\nodepart{two}
				10
				\nodepart{three}
				C
			};
			\node[rectangle split, rectangle split parts=3, rectangle split part align={center,center},
			draw, rectangle split part fill={white!0,white!0,backBlue}, right=0.5 cm of second] (third)
			{
				pants
				\nodepart{two}
				10
				\nodepart{three}
				A
			};
			\node[rectangle split, rectangle split parts=3, rectangle split part align={center,center},
			draw, rectangle split part fill={white!0,white!0,backPurple}, right=0.5 cm of third] (fourth)
			{
				pants
				\nodepart{two}
				20
				\nodepart{three}
				A
			};
			\path[-{Latex[length=2mm,width=2mm]}] (first) edge (second);
			\path[-{Latex[length=2mm,width=2mm]}] (second) edge (third);
			\path[-{Latex[length=2mm,width=2mm]}] (third) edge (fourth);
			\node[rectangle split, rectangle split parts=3, rectangle split part align={right,right}, left=0.2 cm of first] (labels)
			{
				\textbf{product:}
				\nodepart{two}
				\textbf{amount:}
				\nodepart{three}
				\textbf{vendor:}
			};
			\node[align=right, above left=0.1 cm and 0.2 cm of first] (event-label) {\textbf{event:}};
			\node[above=0.1cm of first] (event1) {\eventnumber \label{event:cat-1-ambiguous-match}};
			\node[above=0.1cm of second] (event2) {\eventnumber \label{event:cat-2-perfect-match}};
			\node[above=0.1cm of third] (event3) {\eventnumber \label{event:cat-3-mismatch}};
			\node[above=0.1cm of fourth] (event4) {\eventnumber \label{event:cat-4-no-rule}};
		\end{tikzpicture}
	\end{mdframed}
	\mycaption{Data encoding}{We use the model (top left), the model stream $C_m$, value stream $C_v$ and rule selection stream $C_r$ (top right) to decode the trace's categorical variable \emph{vendor} (bottom).}
	\label{fig:data-encoding-cat-example}
\end{figure}

\subsection{Data encoding}

To encode a given event log with a model of data modification rules,
we encode the variable values at each event.
For each event, we check which rules in the model fire, i.e., have satisfied conditions,
and use the firing rules to encode the data values.
Whenever the model makes ambiguous predictions,
we encode the specific value among all possibilities.
If no rule fires, we choose and encode a value from the whole domain of the target variable.
To ensure a lossless encoding and to handle noisy data, we also encode any errors made by the model.

Conceptually, we split the data encoding into three code streams:
In the rule selection stream $C_r$, we encode which of the rules with matching conditions we choose to encode the current variable value.
Then, we encode in the model stream $C_m$ if the model predicts the correct value.
If not, any value of the target domain is possible.
Whenever the model predicts multiple values, we choose a value by a code in the value stream $C_v$.

We give a toy example of a data encoding in Figure~\ref{fig:data-encoding-cat-example},
which we use to describe how to decode the categorical variable \emph{vendor}.
First, at event~\refevent{event:cat-1-ambiguous-match}, we see that only rule~\refrule{rule:cat-prediction-2} applies.
Thus, we do not need to select a rule by reading from $C_r$.
Next, we find a checkmark as the first symbol in $C_m$, i.e., the model predicts correctly.
However, the rule allows two values, B and C, which we disambiguate by reading C from $C_v$.
Next, at event~\refevent{event:cat-2-perfect-match}, we observe that both rules apply.
Therefore, we check $C_r$ to find that we should use rule~\refrule{rule:cat-prediction-1} whose prediction is correct according to the second symbol in $C_m$.
Further, its prediction is not ambiguous and we get the value C in the trace.
Afterwards, at event~\refevent{event:cat-3-mismatch}, we find that only rule~\refrule{rule:cat-prediction-1} applies but its prediction is incorrect according to the last element in $C_m$.
We obtain the correct value by reading A from $C_v$.
Finally, no rule applies at event~\refevent{event:cat-4-no-rule},
so we neither read from $C_m$ nor from $C_r$.
Instead, we read the last value from $C_v$, A,
by which we have successfully decoded the values for \emph{vendor}.

We compute the length of the data encoding by summing the code lengths in $C_r$, $C_m$ and $C_v$.
Whenever we have to disambiguate multiple firing rules in $C_r$,
we assume all rules in the model are equally important.
We compute the encoded length of $C_r$ by
\begin{equation}
	L(C_r) = \sum_{i=1}^{|C_r|} \log\abs{R_i}\;,
\end{equation}
where $|R_i|$ denotes the set of firing rules at the $i$-event.

When we compute the encoded length of $C_m$, we do not know the probabilities of codes for checkmarks and crosses in advance.
Therefore, we use a prequential plug-in code \cite{dawid_present_1984,rissanen_universal_1984} to compute $L(C_m)$:
We initialize uniform counts for checkmarks and crosses and update counts after each event,
such that we have a valid probability distribution at each step in the encoding.
Asymptotically, this gives an optimal encoded length of $C_m$.
Formally, we have
\begin{equation}
	L(C_m) = \sum_{i = 1}^{|C_m|} \frac{\usg_i C_m[i] + \epsilon}{\usg_i \text{\checkmarkCode} + \usg_i \text{\crossCode} + 2\epsilon}\;,
\end{equation}
where $\usg_i C_m[i]$ denotes how often the $i$-th code in $C_m$ has been used before,
and $\epsilon$ with standard choice $0.5$ is for additive smoothing.

To compute code lengths in $C_v$, we use the conditional empirical probability of values.
Let $u_i$ be the update rule we selected in $C_r$ to encode the $i$-th value in $C_v$.
If no rule fires or we encoded an error in $C_m$, $u_i$ falls back to all possible values in the domain of the target variable.
We formally define
\begin{equation}
	L(C_v) = -\sum_{i=1}^{|C_v|}\log\frac{\freq(C_v[i])}{\sum_{j \in u_i} \freq(j)}\;,
\end{equation}
where $\freq(C_v[i])$ denotes how often value $C_v[i]$ occurs in the data,
and we normalize this by the frequencies of all values $j$ possible according to the update rule $u_i$.
To encode numerical variables, we assume a histogram-based discretization,
by which we can compute all necessary probabilities and code lengths.

Altogether, this gives us a lossless data encoding.

\subsection{Model encoding}
To define the length of the model encoding $L(M)$,
we need to encode the number of rules in the model and all conditions $c$ and update rules $u$ in the model.
Since the number of rules is unbounded,
we use the universal MDL encoding for natural numbers $\LN$~\cite{rissanen:83:integers} that encodes any natural number $x \ge 1$ as
$
	\LN(x) = \log(2.865064) + \log(x) + \log(\log(x)) + \dots,
$
where we sum only the positive terms and the first term ensures that Kraft's inequality holds, i.e., $\LN$ is a lossless encoding.
Denoting the number of modification rules in the model as $|M|$,
which can be zero,
we then obtain the encoded length of the model as
\begin{equation}
	L(M) = L_\mathbb{N}(|M| + 1) + \sum_{(c,u) \in M}L(c) + L(u)\;.
\end{equation}
To encode a condition $c$, we specify its type,
which single variable $v \in V$ is tested by $c$,
and all constants used in the condition.
Formally, we define
\begin{equation}
	L(c) = \log\abs{\{=,\neq,\leq,\geq,\rightarrow\}} + \log\abs{V} + \sum_{\alpha\in c} L(\alpha)\;.
\end{equation}
To encode the value $\alpha \in \dom{v}$ for a categorical variable $v$, we have
\[
L(\alpha) = \log\abs{\dom{v}}\;,
\]
and to encode a real-valued $\alpha$, we consider it in scientific notation $\alpha = k \cdot 10^s$, and have
\[
\LR(\alpha) = 2 + L_\mathbb{N}(|s| + 1) + L_\mathbb{N}(\left\lceil |k| \cdot 10^s \right\rceil + 1)\;,
\]
where we first encode the signs of $k$ and $s$ with 1 bit each,
then the value of $s$,
and finally the value of $k$ up to a user-specified precision of $p$ significant digits \cite{marx_telling_2019}.

To encode an update rule $u$ on a variable $v$,
we first specify the type of $u$, which is one of
$v \in \{\alpha, \beta, \ldots\}$, $v = \alpha$, $v \in [\alpha,\beta]$,
$v = v + \alpha$, $v = v + [\alpha, \beta]$, or $v = \alpha \cdot v$,
for which we need $\log 6$ bits.
Then, we encode which variable $v \in V$ is updated by $u$,
and finally we encode the constants in $u$ the same way we do for conditions.
Formally, we have
\begin{equation}
	L(u) = \log 6 + \log\abs{V} + \sum_{\alpha\in u} L(\alpha)\;.
\end{equation}
This gives us the encoded length of the model $L(M)$.

\subsection{Formal problem definition}
With this, we now have all the ingredients to formally define our problem.

\vspace{0.5em}
\noindent\textbf{Minimal Modification Rules Problem}
\textit{
	Given an event log $D$ with variables $V$,
	find an acyclic model of data modification rules $M$
	that minimizes the total encoding cost $L(D, M) = L(M) + L(D \mid M)$.
}

\vspace{0.5em}
\noindent In practice, it is infeasible to solve this problem optimally due to the potentially large number of acyclic models.
To approximate this number by a lower bound,
we consider the \emph{rule dependency graph} of a model,
which is a directed acyclic graph
with variables as nodes and their dependencies induced by modification rules as edges.
We give a simple example of a rule dependency graph in Figure~\ref{fig:rule-dependency-graph}.
Since each edge requires at least one rule,
there are at least as many models as rule dependency graphs.
According to Rodionov \cite{rodionov_number_1992},
the number of acyclic graphs with $n$ nodes and up to $m$ edges is
\[
	A(n,m) = \sum_{i=1}^n \sum_{j=0}^m (-1)^{i-1} \binom{n}{i} \binom{i(n-i)}{m-j} A(n-i,j)\;,
\]
with $A(1,\cdot):=1$.
Because $A(n,m)$ grows exponentially in $n$ and $m$ \cite[p.\  1186]{cormen:2009:introduction},
the number of acyclic models grows exponentially in the number of variables $|V|$ and the number of modification rules in the model $|M|$.

Furthermore, our search space has no trivial structure such as submodularity or monotonicity,
which we could exploit to find an optimal solution in feasible time.
We give counterexamples for both properties in the supplementary material.
Hence, we resort to heuristics.

\begin{figure}
	\begin{mdframed}
		\centering
		\setlength\tabcolsep{1pt}
		\begin{tabular}{ll}
			\textbf{IF} \textit{product} $=$ pants & \textbf{THEN} \textit{vendor} $=$ A\\
			\textbf{IF} \textit{product} $=$ bag & \textbf{THEN} \textit{amount} $=$ 20\\
		\end{tabular}
		\tikzsetnextfilename{rule-dependency-graph}
		\begin{tikzpicture}[node distance=2.5cm,auto]
			\usetikzlibrary{shapes.geometric}
			\usetikzlibrary{arrows.meta}
			\node[draw,ellipse] (am) {\textit{amount}};
			\node[draw,ellipse] (prod) [right of=am] {\textit{product}};
			\node[draw,ellipse] (vend) [right of=prod] {\textit{vendor}};
			\path[-{Latex[length=2mm,width=2mm]}] (prod) edge (vend);
			\path[-{Latex[length=2mm,width=2mm]}] (prod) edge (am);
		\end{tikzpicture}
	\end{mdframed}
	\mycaption{Rule dependency graph}{We represent the variable dependencies of the model (top) as a graph (bottom) where ovals represent variables and arrows show whether and in which direction a modification rule induces a dependency between variables.}
	\label{fig:rule-dependency-graph}
\end{figure}

%% file: algorithm.tex
\section{The \ourmethod algorithm}

To efficiently discover good sets of data modification rules in practice,
we prune the exponentially sized search space with
a quickly computable estimate of our score that avoids repetitively
passing the whole event log, and we introduce a greedy search.

\subsection{Estimating the MDL score}

Computing the MDL score $L(D, M)$ requires a pass over all events in the event log,
because we must check for each event, which of the rules in the model fire.
To avoid iterating over the event log each time we evaluate adding a new rule to the model,
we prune the large set of potential candidate rules by a quickly computable estimate $\widehat{L}(D, M)$.
We optimistically assume that a newly generated rule is the only rule in the model which predicts its target variable,
such that we do not need to update $C_r$ or $C_m$.

\begin{algorithm}[tb!]
	\caption{Estimate $L(C_v)$}\label{alg:estimate}
	\KwIn{Rule $(c, u)$}
	\KwOut{$\widehat{L}(C_v \mid c, u)$}
	$\widehat{L}(C_v \mid c, u) \gets 0$\;
	$b \gets \supp(c)$\;\label{line:estimate-compute-support}
	\ForAll{$j \in u$ ordered by increasing $\freq(j)$}{
		$\Delta b \gets \min\{b, \freq(j)\}$\;\label{line:estimate-delta-b}
		$\widehat{L}(C_v~|~c, u) \gets \widehat{L}(C_v~|~c, u) - \Delta b \cdot \log \frac{\freq(j)}{\sum_{i \in u}\freq(i)}$\;\label{line:estimate-add-codes}
		$b \gets b - \Delta b$\;\label{estimate:update-b}
		}
	\Return{$\widehat{L}(C_v \mid c, u)$}\;
\end{algorithm}

By assuming independence between rules,
we can independently estimate the contribution of a single rule with condition $c$ and update rule $u$
to $L(C_v)$.
We give the pseudocode for estimating $L(C_v\mid c,u)$ as Algorithm~\ref{alg:estimate}.
First, we compute the support $\supp(c)$, i.e.,
at how many events $c$ fires (ln.\ \ref*{line:estimate-compute-support}).
For each value $j$ predicted by $u$,
we compute how many codes we must add to $C_v$,
which is the minimum of the remaining events to cover, $b$,
and the frequency $\freq(j)$ of $j$ (ln.\ \ref*{line:estimate-delta-b}).
We compute the length of all these codes and add it to our estimate (ln.\ \ref*{line:estimate-add-codes}).
At the end of each iteration, we update how many codes we still must add to $C_v$ (ln.\ \ref*{estimate:update-b}).

While computing the support requires a pass over the dataset,
we only need a single pass when creating the candidate.
By assuming independence between rules,
we do not need to update the estimated code lengths of all candidate rules,
every time we add another rule to the model.
Using $\widehat{L}(D, M)$ we can prune out rules with high encoding costs,
and thus avoid computing $L(D, M)$ for those.
Next, we use $\widehat{L}(D, M)$ and $L(D, M)$ to find good modification rules for a given event log.

\subsection{Finding Good Modification Rules}
To reduce the search space, we let the domain experts control how much time they want to invest in model search,
and introduce two hyperparameters for a greedy search.
Instead of generating conditions with all possible combinations of variables, signs and values,
we only generate the $\paramNc$ most frequently observed combinations of variables and values in the dataset for each sign of the set $\{=,\neq,\leq,\geq,\rightarrow\}$.
For a given condition, instead of generating update rules with all possible combinations of variables and values,
we only generate the $\paramNu$ most frequent combinations of variables and values in the dataset for each type of update rule.

\begin{algorithm}[tb!]
	\caption{\ourmethod}\label{alg:moody}
	\KwIn{event log $D$ with variables $V$}
	\KwOut{model of data modification rules $M$}
	$M \gets \emptyset$\; \label{line:empty-model}
	\Do{$M$ was extended in the last iteration\label{line:outer-loop}}{
		\ForAll{$v \in V$\label{line:variable-loop}}{
			$Q \gets$ priority queue of rules $r$ predicting $v$ ordered by $\widehat{L}(D, M \cup \{r\})$\; \label{line:priority-queue}
			$r^* \gets \emptyset$\;
			\While{$Q \neq \emptyset~\AND~\widehat{L}(D, M \cup \{\text{top of }Q\}) < L(D, M \cup \{r^*\})$\label{line:search-loop}}{
				$r' \gets $ pop element from $Q$\;
				$r^* \gets \argmin_{r\in\{r^*,r'\}} L(D, M \cup \{r\})$
				\label{line:search-loop-end}}
			\If{$L(D, M \cup \{r^*\}) < L(D, M)$\label{line:check-improvement}}{
				$M \gets M \cup \{r^*\}$\; \label{line:add-best-rule}
			}
			\label{line:variable-loop-end}}
		\label{line:outer-loop-end}}
	\Return{$M$}\; \label{line:model-return}
\end{algorithm}

We provide the pseudocode of our greedy search \ourmethod as Algorithm~\ref{alg:moody}.
Our search starts with an empty model (\refline{line:empty-model}).
We iteratively extend this model by modification rules for all target variables (\refline{line:variable-loop}).
Since computing $L(D, M)$ needs a pass over the whole dataset,
we manage the candidate rules in a priority queue sorted by an estimate of our score $\widehat{L}(D, M)$ (\refline{line:priority-queue}),
such that we check promising rules early.
Next, we search the candidates from most promising to least promising and compute their actual encoded length $L$ (ln.\ \ref*{line:search-loop}-\ref*{line:search-loop-end}).
To not waste computation time for computing $L$ on inferior rules,
we perform this search as long as the estimate $\widehat{L}$ is better than the best actual code length $L$ that we have seen so far (\refline{line:search-loop}).
After evaluating the candidates, we only add the best candidate to our model if it reduces the total encoded length (ln.\ \ref*{line:check-improvement}-\ref*{line:add-best-rule}).
Finally, we end when no candidate for any target variable could improve our score (\refline{line:outer-loop}) and return the resulting model (\refline{line:model-return}).

In the worst case, all generated candidates improve our MDL score.
Since the number of candidates grows linearly with $\paramNc$ and $\paramNu$,
the outer loop of \ourmethod grows linearly with $\paramNc$ and $\paramNu$.
In the worst case, our estimate does not prune any candidate,
and we must compute our score for all of the $O(\paramNc \cdot \paramNu)$ candidates.
Since we loop over all variables, and computing our score requires a pass over the whole dataset,
the runtime complexity of \ourmethod is $O\left((\paramNc \cdot \paramNu)^2 \cdot |V| \cdot |D|\right)$.

%% file: experiments.tex
\section{Experiments}
\label{sec:exps}
In this section, we empirically evaluate \ourmethod on both synthetic and real-world data.
When we defined our MDL score, we assumed discretization of numerical variables.
In our prototype implementation, we discretize numerical values into variable-width histograms.
For efficiency, we determine the histogram boundaries by percentiles.
We use 50 bins in all our experiments.

We run all our experiments in a Docker-based environment on a Linux server with an Intel\textsuperscript{\textregistered} Xeon\textsuperscript{\textregistered} Gold 6244 CPU.
In all experiments, we observe 16 GB of RAM suffice.
As a simple baseline, we consider the empty model $M = \emptyset$.
In addition, we learn if-then-else rules using the rule-based classifier \turs \cite{yang:2022:turs},
and using the subgroup discovery method \ssdpp \cite{proenca:2022:ssdpp}.
To ensure reproducibility of all our results, we provide code, data, and further details in the supplementary material.

\begin{figure}
	\tikzsetnextfilename{moody_hyperparameter_tuning}
	\begin{tikzpicture}
		\begin{axis}[
			name=ax1,
			eda line,
			height=4cm,
			width=4.5cm,
			ylabel={\fonescore},
			y label style={at={(axis description cs:-0.20,0.5)}, anchor=south, font=\scriptsize},
			xlabel={Hyperparameter $\paramNc$},
			ymin=0.0,
			ymax=1.0,
			yticklabels={},
			extra y ticks={0.0,0.1,0.2,0.3,0.4,0.5,0.6,0.7,0.8,0.9,1.0},
			extra y tick labels={0.0,\empty,0.2,\empty,0.4,\empty,0.6,\empty,0.8,\empty,1.0},
			xmin=2.0,
			xmax=70.0,
			xticklabels={2,10,30,50,70},
			xtick={2,10,30,50,70},
			extra x ticks={},
			]
			\addplot+[discard if not={noiseProportion}{0.1}, error bars/.cd, y dir=both, y explicit] table [col sep=comma, x=conditionCount, y=midIqr, y error minus=relative lowIqr, y error plus=relative highIqr] {expres/prediction-f1-condition-count.csv};
			\addplot+[discard if not={noiseProportion}{0.2}, error bars/.cd, y dir=both, y explicit] table [col sep=comma, x=conditionCount, y=midIqr, y error minus=relative lowIqr, y error plus=relative highIqr] {expres/prediction-f1-condition-count.csv};
		\end{axis}
		\begin{axis}[
			name=ax2,
			at={(ax1.south east)},
	    	xshift=-0.76cm,
			eda line,
			legend pos=north east,
			xshift=2cm,
			height=4cm,
			width=4.5cm,
			ylabel={RMSE},
			xlabel={Hyperparameter $\paramNc$},
			ymin=0.0,
			ymax=30.0,
			yticklabels={},
			extra y ticks={0,5,10,15,20,25,30},
			extra y tick labels={0,5,10,15,20,25,30},
			xmin=2.0,
			xmax=70.0,
			xticklabels={2,10,30,50,70},
			xtick={2,10,30,50,70},
			extra x ticks={},
			y label style={at={(axis description cs:-0.19,0.5)}, anchor=south, font=\scriptsize},
			legend style={
				legend columns=-1,
				at={(-0.3,1.3)},anchor=north
			},
			]
			\addplot+[discard if not={noiseProportion}{0.1}, error bars/.cd, y dir=both, y explicit] table [col sep=comma, x=conditionCount, y=midIqr, y error minus=relative lowIqr, y error plus=relative highIqr] {expres/prediction-rmse-condition-count.csv};
			\addlegendentry{$10\%$ swap noise}
			\addplot+[discard if not={noiseProportion}{0.2}, error bars/.cd, y dir=both, y explicit] table [col sep=comma, x=conditionCount, y=midIqr, y error minus=relative lowIqr, y error plus=relative highIqr] {expres/prediction-rmse-condition-count.csv};
			\addlegendentry{$20\%$ swap noise}
		\end{axis}
	\end{tikzpicture}
	\mycaption{Choosing $\paramNc$}{
		Median \fonescore on categorical variables (left) and median RMSE on numerical variables (right)
		for different values of \ourmethod{}'s hyperparameter $\paramNc$ for $10\%$ and $20\%$ swap noise in the training set.
		Error bars show interquartile ranges.
	}
	\label{fig:condition-count}
\end{figure}

\subsection{Synthetic event logs}
To control data properties such as noise,
we first experiment on synthetic event logs,
such that we know the generating ground-truth rules.
We randomly generate ten independent ground-truth models in our modeling language,
where each model contains five rules, two categorical variables and two numerical variables.
Since \ssdpp cannot model sequential dependencies $v: \alpha \rightarrow \beta$,
we only create models with conditions $v \le \alpha$, $v \ge \alpha$ and $v = \alpha$.
From these ground-truth models, we generate event logs with $|D| = 2000$ events.
To test noise-robustness, we add different amounts of noise,
where we randomly swap values of variables.
10\% swap noise means that for each variable in the dataset,
we randomly swap 10\% of its values.

\turs in contrast to \ssdpp does only discover rules for categorical target variables, and cannot predict numerical values.
Therefore, we separately evaluate on models that only predict categorical variables
and on models that only predict numerical variables.
However, in both setups, we generate conditions containing categorical and numerical variables.

\paragraph{Choosing Hyperparameters}
Such that the user can control runtime by reducing the search space,
we introduced hyperparameters $\paramNc$ and $\paramNu$ when we proposed \ourmethod.
For efficiency, we only search for the most compressing update rule given a condition and set $\paramNu$ to 1.
To evaluate the accuracy of a set of rules, we compute its median \fonescore on predicting categorical variables,
and its median root mean squared error (RMSE) in predicting numerical variables.
We report the influence of $\paramNc$ on prediction accuracy in Figure~\ref{fig:condition-count}.
We see that the prediction accuracy drops for only very small $\paramNc$,
and is relatively constant for larger values.
This means, the conditions with the highest support in the dataset tend to give
the best compression and accuracy.
To have a safety margin on unknown data, we set $\paramNc$ to 50.

\begin{figure*}[tb!]
	\tikzsetnextfilename{moody_predictions_on_synthetic_data}
	\begin{tikzpicture}
		\begin{axis}[
			name=ax1,
			eda line,
			width=5.8cm,
			height=4cm,
			ylabel={Test \fonescore},
			y label style={at={(axis description cs:-0.14,0.5)}, anchor=south, font=\scriptsize},
			xlabel={Noise proportion},
			ymin=0.0,
			ymax=1.0,
			yticklabels={},
			extra y ticks={0.0,0.1,0.2,0.3,0.4,0.5,0.6,0.7,0.8,0.9,1.0},
			extra y tick labels={0.0,\empty,0.2,\empty,0.4,\empty,0.6,\empty,0.8,\empty,1.0},
			]
			\addplot+[discard if not={method}{my}, error bars/.cd, y dir=both, y explicit] table [col sep=comma, x=noiseProportion, y=midIqr, y error minus=relative lowIqr, y error plus=relative highIqr] {expres/prediction-f1-noise-vs-ssdpp.csv};
			\addplot+[discard if not={method}{turs}, error bars/.cd, y dir=both, y explicit] table [col sep=comma, x=noiseProportion, y=midIqr, y error minus=relative lowIqr, y error plus=relative highIqr] {expres/prediction-f1-turs-mixed.csv};
			\addplot+[discard if not={method}{ssdpp}, error bars/.cd, y dir=both, y explicit] table [col sep=comma, x=noiseProportion, y=midIqr, y error minus=relative lowIqr, y error plus=relative highIqr] {expres/prediction-f1-noise-vs-ssdpp.csv};
			\addplot coordinates {(0.0, 0.20359027898464238) (0.3, 0.20359027898464238)};
		\end{axis}
		\begin{axis}[
			name=ax2,
			at={(ax1.south east)},
	    	xshift=1.5cm,
			eda line,
			width=5.8cm,
			height=4cm,
			ylabel={Test RMSE},
			y label style={at={(axis description cs:-0.13,0.5)}, anchor=south, font=\scriptsize},
			xlabel={Noise proportion},
			ymin=0.0,
			ymax=30.0,
			extra y ticks={5,15,25},
	        extra y tick labels={5,15,25},
			]
			\addplot+[discard if not={method}{my}, error bars/.cd, y dir=both, y explicit] table [col sep=comma, x=noiseProportion, y=midIqr, y error minus=relative lowIqr, y error plus=relative highIqr] {expres/prediction-rmse-noise-vs-ssdpp.csv};
			\pgfplotsset{cycle list shift=1}
			\addplot+[discard if not={method}{ssdpp}, error bars/.cd, y dir=both, y explicit] table [col sep=comma, x=noiseProportion, y=midIqr, y error minus=relative lowIqr, y error plus=relative highIqr] {expres/prediction-rmse-noise-vs-ssdpp.csv};
			\pgfplotsset{cycle list shift=1}
			\addplot coordinates {(0.0, 22.087389912697557) (0.3, 22.087389912697557)};
		\end{axis}
		\begin{axis}[
			name=ax3,
			at={(ax2.south east)},
	    	xshift=1.5cm,
			eda line,
			width=5.8cm,
			height=4cm,
			ylabel={No.\ of rule terms},
			y label style={at={(axis description cs:-0.14,0.5)}, anchor=south, font=\scriptsize},
			xlabel={Noise proportion},
			ymax=100.0,
			ymin=0,
	        extra y ticks={10,30,50,70,90},
	        extra y tick labels={},
			legend style={
				legend columns=-1,
				at={(-0.8,1.3)},anchor=north
			},
			]
			\addplot+[discard if not={method}{my}, error bars/.cd, y dir=both, y explicit] table [col sep=comma, x=noiseProportion, y=midIqr, y error minus=relative lowIqr, y error plus=relative highIqr] {expres/complexity-noise-vs-ssdpp.csv};
			\addlegendentry{\ourmethod}
			\addplot+[discard if not={method}{turs}, error bars/.cd, y dir=both, y explicit] table [col sep=comma, x=noiseProportion, y=midIqr, y error minus=relative lowIqr, y error plus=relative highIqr] {expres/complexity-turs-mixed.csv};
			\addlegendentry{\turs}
			\addplot+[discard if not={method}{ssdpp}, error bars/.cd, y dir=both, y explicit] table [col sep=comma, x=noiseProportion, y=midIqr, y error minus=relative lowIqr, y error plus=relative highIqr] {expres/complexity-noise-vs-ssdpp.csv};
			\addlegendentry{\ssdpp}
			\addplot+ coordinates {(0.0, 0.0) (0.3, 0.0)};
			\addlegendentry{empty model}
			\addplot+[color=black,densely dotted] coordinates {(0.0, 10.0) (0.3, 10.0)};
			\addlegendentry{ground-truth model}
		\end{axis}
	\end{tikzpicture}
	\mycaption{\ourmethod predicts well under reasonable amounts of noise}{
		Median \fonescores on categorical variables (left, higher is better),
		median root mean squared errors on numerical variables (center, lower is better)
		and number of rule terms in the discovered models (right, lower is less complex)
		at different noise levels
		for \ourmethod, \ssdpp{} and \turs{}. Error bars indicate interquartile ranges.
	}
	\label{fig:synth_data_results}
\end{figure*}

\paragraph{Results on categorical variables}
Next, we compare results on synthetic data with different amounts of swap noise for \ourmethod against all baselines in Figure~\ref{fig:synth_data_results}.
We see in the left plot that \ourmethod achieves by a wide margin the highest \fonescore on categorical variables
for data with 0\% and 10\% noise.
Up to 20\% noise, \ourmethod shows good noise-robustness and still has the highest median \fonescore.
For 30\% noise, the \fonescore of \ourmethod drops down to the \fonescore of the empty model.
We note that 30\% noise may sound low, but swapping 30\% of the values for each variable
in the dataset accumulates to a much higher noise-ratio.
Hence, \ourmethod predicts well under reasonable amounts of noise.

\paragraph{Results on numerical variables}
We see similar results on predicting values of numerical variables in the center plot of Figure~\ref{fig:synth_data_results}.
Since \turs cannot predict numerical variables, we compare \ourmethod to \ssdpp and the empty model.
\ourmethod shows by far the smallest test root mean squared error (RMSE) for training data with up to 20\% noise.
For 30\% noise, \ourmethod's RMSE converges to the RMSE of the empty model.
We again note swapping 30\% of the values for each variable
in the dataset accumulates to a much higher noise-ratio
than we expect in any real-world event log.
Hence, \ourmethod predicts well under reasonable amounts of noise.

\paragraph{Model complexity}
Not only does \ourmethod give the best prediction results under reasonable amounts of noise.
It also discovers the rule sets with the lowest total number of rule terms as we show in the right plot of Figure~\ref{fig:synth_data_results}.
Both \turs and \ssdpp find rule sets with a significantly higher number of rule terms.
We see \ourmethod's models have similar complexity compared to the ground-truth models for data with low amounts of noise.
If the noise level increases, \ourmethod converges to the empty model.
This means, \ourmethod is robust against finding spurious rules on noisy data.

\begin{figure}
	\tikzsetnextfilename{moody_sample_complexity}
	\begin{tikzpicture}
		\begin{axis}[
			name=ax1,
			eda line,
			height=4cm,
			width=4.5cm,
			ylabel={Test \fonescore},
			y label style={at={(axis description cs:-0.20,0.5)}, anchor=south, font=\scriptsize},
			xlabel={No.\ of training events},
			ymin=0.0,
			ymax=1.0,
			yticklabels={},
			extra y ticks={0.0,0.1,0.2,0.3,0.4,0.5,0.6,0.7,0.8,0.9,1.0},
			extra y tick labels={0.0,\empty,0.2,\empty,0.4,\empty,0.6,\empty,0.8,\empty,1.0},
			xticklabels={},
			extra x ticks={250,500,750,1000,1250,1500},
			extra x tick labels={250,500,\empty,1000,\empty,1500},
			legend style={
				legend columns=-1,
				at={(1.25,1.3)},anchor=north
			},
			]
			\addplot+[discard if not={method}{my}, error bars/.cd, y dir=both, y explicit] table [col sep=comma, x=logLength, y=midIqr, y error minus=relative lowIqr, y error plus=relative highIqr] {expres/scaling-f1-score.csv};
			\addlegendentry{\ourmethod}
			\addplot+[discard if not={method}{turs}, error bars/.cd, y dir=both, y explicit] table [col sep=comma, x=logLength, y=midIqr, y error minus=relative lowIqr, y error plus=relative highIqr] {expres/scaling-f1-score.csv};
			\addlegendentry{\turs}
			\addplot+[discard if not={method}{ssdpp}, error bars/.cd, y dir=both, y explicit] table [col sep=comma, x=logLength, y=midIqr, y error minus=relative lowIqr, y error plus=relative highIqr] {expres/scaling-f1-score.csv};
			\addlegendentry{\ssdpp}
			\addplot+ coordinates {(250.0, 0.26366860257685154) (1500.0, 0.26366860257685154)};
			\addlegendentry{empty model}
		\end{axis}
		\begin{axis}[
			name=ax2,
			at={(ax1.south east)},
	    	xshift=-0.76cm,
			eda line,
			legend pos=north east,
			xshift=2cm,
			height=4cm,
			width=4.5cm,
			ylabel={Runtime ${[s]}$},
			xlabel={No.\ of training events},
			ymin=0.0,
			ymax=60.0,
			yticklabels={},
			extra y ticks={0.0,10.0,20.0,30.0,40.0,50.0,60.0},
			extra y tick labels={0,10,20,30,40,50,60},
			xticklabels={},
			extra x ticks={250,500,750,1000,1250,1500},
			extra x tick labels={250,500,\empty,1000,\empty,1500},
			y label style={at={(axis description cs:-0.19,0.5)}, anchor=south, font=\scriptsize},
			legend style={
				legend columns=-1,
				at={(-0.3,1.3)},anchor=north
			},
			]
			\addplot+[discard if not={method}{my}, error bars/.cd, y dir=both, y explicit] table [col sep=comma, x=logLength, y=midIqr, y error minus=relative lowIqr, y error plus=relative highIqr] {expres/scaling-runtime.csv};
			\addplot+[discard if not={method}{turs}, error bars/.cd, y dir=both, y explicit] table [col sep=comma, x=logLength, y=midIqr, y error minus=relative lowIqr, y error plus=relative highIqr] {expres/scaling-runtime.csv};
			\addplot+[discard if not={method}{ssdpp}, error bars/.cd, y dir=both, y explicit] table [col sep=comma, x=logLength, y=midIqr, y error minus=relative lowIqr, y error plus=relative highIqr] {expres/scaling-runtime.csv};
		\end{axis}
	\end{tikzpicture}
	\mycaption{\ourmethod shows low sample complexity and scales well}{
		Median test \fonescore on categorical variables (left)
		and median runtime (right)
		dependent on the number of training events for \ourmethod, \turs and \ssdpp.
		Error bars show interquartile ranges.
	}
	\label{fig:moody-sample-complexity}
\end{figure}

\paragraph{Sample complexity}

To empirically evaluate sample complexity,
we compute $\fonescores$ on the test set dependent on
the number of events in the training set.
For a realistic setup, we add 10\% swap noise to all training sets.
We report results for \ourmethod, \turs, \ssdpp and the empty model in the left plot of Figure~\ref{fig:moody-sample-complexity}.
We see \ourmethod predicts better the more training data is available,
while all baselines do not improve with more training data.
Already with 500 training events, \ourmethod shows higher median $\fonescore$ than the baselines.

\paragraph{Runtime}

We report wall-clock runtime for single-threaded execution dependent on the number of training events in the right plot of Figure~\ref{fig:moody-sample-complexity}.
As we expect by our theoretical runtime analysis,
we see that \ourmethod scales well and shows a growth of runtime linear to the number of training events.
While \ssdpp shows a constantly fast, almost zero runtime,
\ourmethod still finishes within reasonable time and is significantly faster than \turs.

\subsection{Real-world event logs}

\begin{table}[tb!]
	\small
	\centering

	\begin{tabular}{l rrrrr}
		\toprule
		Data & $\nrOfTraces$ & $\nrOfEvents$ & $|\Omega|$ & $\abs{\setOfCategoricalVariables}$ & $\abs{\setOfNumericalVariables}$ \\
		\midrule
		\sepsis       & 782   & 15214  & 18 & 25 & 7\\
		\trafficfines & 30074 & 112245 & 11 & 4  & 5\\
		\bottomrule
	\end{tabular}
	\caption{
        [\textbf{Real-world event logs statistics}]
        Number of traces $\nrOfTraces$,
		number of events $\nrOfEvents$,
        number of different activities $|\Omega|$,
		number of categorical variables $\abs{\setOfCategoricalVariables}$,
		and number of categorical variables $\abs{\setOfNumericalVariables}$
		for the \sepsis and \trafficfines real-world datasets.
    }
	\label{tab:moody_real_world_datasets}
\end{table}

Next, we evaluate on two publicly available real-world event logs,
for which we give the base statistics in Table~\ref{tab:moody_real_world_datasets}.
The first one, \emph{\sepsis} \cite{mannhardt:2016:sepsis}, contains event traces from
treating Sepsis patients in a Dutch hospital.
The second one, \emph{\trafficfines} \cite{deleoni:2015:trafficfines}, \cite[p. 20]{mannhardt:2016:conformance} is an event log
of handling road-traffic fines by the police of an Italian city.
To reduce runtime, we randomly sample 20\% of the original 150370 traces in the \trafficfines event log.
Furthermore, we parallelize candidate generation and candidate evaluation of \ourmethod
on twelve CPU cores.

First, we look at the insight we gain from some exemplary rules we find with \ourmethod on these event logs.
Then, we evaluate how well the discovered rules generalize to unseen test data.

\begin{figure}
	\begin{mdframed}
		\centering
		\setlength\tabcolsep{1pt}
		\begin{tabular}{ll}
			\textbf{IF} \textit{group} $=$ C & \textbf{THEN} \textit{activity} $=$ ER Triage\\
			\textbf{IF} \textit{group} $=$ E & \textbf{THEN} \textit{activity} $=$ Release A\\
			\textbf{IF} \textit{group} $=$ W & \textbf{THEN} \textit{activity} $=$ Admission IC\\
			\textbf{IF} \textit{group} $=$ P & \textbf{THEN} \textit{activity} $=$ Admission IC\\
			\textbf{IF} \textit{group} $=$ F & \textbf{THEN} \textit{activity} $=$ Admission NC\\
			\textbf{IF} \textit{group} $=$ O & \textbf{THEN} \textit{activity} $=$ Admission NC\\
		\end{tabular}
	\end{mdframed}
	\mycaption{Responsibility rules for the Sepsis event log}{Different groups in the hospital are associated with different activities for handling sepsis patients.}
	\label{fig:sepsis-responsibility-rules}
\end{figure}

\begin{figure}
	\begin{mdframed}
		\centering
		\setlength\tabcolsep{2pt} 
		\begin{tabular}{ll}
			\textbf{IF} \textit{Leukocytes} $\le$ 4.7 & \textbf{THEN} \textit{Infusion} $=$ True\\
			\textbf{IF} \textit{Leukocytes} $\le$ 2.8 & \textbf{THEN} \textit{Diagnose} $=$ GB\\
			\textbf{IF} \textit{Leukocytes} $=$ 7.8 & \textbf{THEN} \textit{Diagnose} $=$ AA
		\end{tabular}
	\end{mdframed}
	\mycaption{Leukocytes rules for the Sepsis event log}{Lower values for leukocytes (white blood cells) are associated with certain diagnoses and infusions.}
	\label{fig:sepsis-leukocytes-rules}
\end{figure}

\paragraph{\sepsis rules}
On the \sepsis dataset, \ourmethod discovers in total 82 rules with an approximate runtime of two hours.
23 of these rules express a correlation between the \emph{group} attribute
and the activity of an event, for which we show a subset in Figure~\ref{fig:sepsis-responsibility-rules}.
The rules found by \ourmethod indicate that certain groups in the hospital are specialized in certain activities.

Furthermore, \ourmethod finds rules, where leukocytes measurements imply the diagnosis and the treatment of sepsis, as we show in Figure~\ref{fig:sepsis-leukocytes-rules}.
These rules enable to ask targeted questions to domain experts and thus can be a valuable start to gain insight into this process.

\begin{figure}
	\begin{mdframed}
		\centering
		\setlength\tabcolsep{2pt} 
		\begin{tabular}{ll}
			\textbf{IF} \textit{amount} $\le$ 68.77 & \textbf{THEN} \textit{points} $=$ 0.0\\
			\textbf{IF} 143.00 $\le$ \textit{amount} & \textbf{THEN} \textit{points} $\in$ [0.0, 10.0]\\
			\textbf{IF} 131.00 $\le$ \textit{amount} & \textbf{THEN} \textit{points} $=$ 0.0\\
			\textbf{IF} \textit{amount} $=$ 80.00 & \textbf{THEN} \textit{points} $\in$ [0.0, 2.0]\\
		\end{tabular}
	\end{mdframed}
	\mycaption{\trafficfines rules}{
		\ourmethod finds multiple relationships between the fine \textit{amount}
		and the \textit{points} deducted from
		the offender's driving license.
	}
	\label{fig:traffic-offence-point-relationship}
\end{figure}

\paragraph{\trafficfines rules}
On the \trafficfines dataset, \ourmethod discovers in total 117 rules with an approximate runtime of 4.5 hours.
While we would expect that a higher fine \emph{amount} correlates with a high number of \emph{points}
deducted from the offender's driving license,
the rules found by \ourmethod contradict this intuition, as we show in Figure~\ref{fig:traffic-offence-point-relationship}.
A closer look on the dataset indeed confirms that there is no monotonic relationship
between these two attributes.
Finding counter-intuitive but data-supported rules like these gives valuable insight into the underlying process.

\begin{figure}
	\tikzsetnextfilename{moody_generalization}
	\begin{tikzpicture}
		\begin{axis}[
			name=ax1,
			eda line,
			height=4cm,
			width=4.5cm,
			ylabel={\fonescore},
			y label style={at={(axis description cs:-0.20,0.5)}, anchor=south, font=\scriptsize},
			xlabel={Rule},
			ymin=0.0,
			ymax=1.0,
			yticklabels={},
			extra y ticks={0.0,0.1,0.2,0.3,0.4,0.5,0.6,0.7,0.8,0.9,1.0},
			extra y tick labels={0.0,\empty,0.2,\empty,0.4,\empty,0.6,\empty,0.8,\empty,1.0}
			]
			\addplot+[discard if not={dataset}{training}] table [col sep=comma, x=ruleNumber, y=f1Score] {expres/trafficFinesGeneralizationCategorical.csv};
			\addplot+[discard if not={dataset}{test}] table [col sep=comma, x=ruleNumber, y=f1Score] {expres/trafficFinesGeneralizationCategorical.csv};
		\end{axis}
		\begin{axis}[
			name=ax2,
			at={(ax1.south east)},
	    	xshift=-0.76cm,
			eda line,
			legend pos=north east,
			xshift=2cm,
			height=4cm,
			width=4.5cm,
			ylabel={RMSE},
			xlabel={Rule},
			ymax=400,
			xmax=100,
			y label style={at={(axis description cs:-0.23,0.5)}, anchor=south, font=\scriptsize},
			legend style={
				legend columns=-1,
				at={(-0.3,1.3)},anchor=north
			},
			]
			\addplot+[discard if not={dataset}{training}] table [col sep=comma, x=ruleNumber, y=RMSE] {expres/trafficFinesGeneralizationNumerical.csv};
			\addlegendentry{train}
			\addplot+[discard if not={dataset}{test}] table [col sep=comma, x=ruleNumber, y=RMSE] {expres/trafficFinesGeneralizationNumerical.csv};
			\addlegendentry{test}
		\end{axis}
	\end{tikzpicture}
	\mycaption{\ourmethod generalizes well}{
		Train and test \fonescore on categorical variables (left)
		and train and test root mean squared error on numerical variables (right)
		for each rule found by \ourmethod on the \trafficfines log.

	}
	\label{fig:generalization}
\end{figure}

\paragraph{Generalization}
Finally, we evaluate how well the rules discovered by \ourmethod generalize to unseen data.
To this end, we split the \trafficfines event log into a training set and a test set with a distinct 20\% of traces each.
Then, we compare the \fonescore respectively RMSE on the training set and the test set for each rule,
which \ourmethod discovers on the training set.
We show results in Figure~\ref{fig:generalization}.
As we see, most of the rules have a low prediction error on both sets.
The gap between training and test performance is small,
which means that \ourmethod finds well-generalizing rules.

%% file: discussion.tex
\section{Discussion}
\label{sec:discussion}

In our experiments, \ourmethod does not only discover simple and thus interpretable rules
to unveil the data modifications within a process.
It also finds rules that accurately predicts the data values,
and is more robust to sensible amounts of noise than all baselines.

Yet, we see multiple interesting research directions to improve \ourmethod.
First, extending the modeling language of \ourmethod would allow understanding data modifications of very complex processes.
For example, this involves rules with complex conditions joined by \emph{(and)} and \emph{(or)}.
We see adapting our MDL score to such rules is relatively easy.

A more complex modeling language implies an even larger search space, and thus we
see the need to improve the runtime efficiency of \ourmethod.
As in many MDL-based methods, the bottleneck of \ourmethod is discrete combinatorial search.
Hence, we see an approach for mining data modification rules
based on differentiable pattern set mining \cite{fischer:2021:binaps} as a promising future direction.

Furthermore, we would like to put a stronger focus on causality of the discovered rules.
To this end, we may use well-defined measures for the causal effect of a rule \cite{budhathoki:21:causalrules}.
Alternatively, we would like to examine the link between causality and two-part MDL codes
in terms of algorithmic independence \cite{marx:2022:formally},
and how to use this during search for causal data modification rules.

Last but not least, we also see many interesting applications for \ourmethod.
As the most compressing rules found by \ourmethod define normal behavior,
it would be interesting to use them for anomaly detection \cite{nolle:2018:binet}.
Since the behavior of real-world processes usually changes over time,
we see \ourmethod could help to identify and understand concept drift \cite{bose:2013:concept-drift,sato:2021:concept-drift}.
Finally, predicting data attributes with \ourmethod may be used
in the simulation of process behavior for process optimization \cite{hlupic:1998:business}.

%% file: conclusion.tex
\section{Conclusion}\label{sec:conclusion}

We studied the hitherto largely neglected problem of discovering accurate
yet concise and interpretable rules how event attribute data changes in a business process.
We formalized the problem in terms of the Minimum Description Length (MDL) principle,
by which we choose the model with the best lossless description of the data.
To efficiently search for rule models in practice, we proposed our greedy method \ourmethod.
Through extensive experiments on both synthetic and real-world data,
we showed \ourmethod indeed discovers succinct and interpretable if-then rules,
needs little data for accurate discovery, is robust to sensible amounts of noise,
and thus gives valuable insight into data modifications.

Besides applying \ourmethod on downstream tasks such as anomaly detection,
concept drift detection and simulation,
future work involves extending the rule language of \ourmethod
to model more complex conditions for data changes,
and runtime optimizations to enable search for such complex rules in feasible time.
\balance

%% file: appendix.tex
\clearpage
\section{Appendix}
\label{sec:apx}

In this section, we give additional details of \ourmethod and our empirical evaluation,
which we could not include into the main paper.

\subsection{Submodularity of our MDL score}

\begin{figure}
	\begin{mdframed}
		\centering
		\textbf{Rules:}\\
		\vspace{0.4em}
		\setlength\tabcolsep{2pt} 
		\begin{tabular}{rrl}
			\rulenumber \label{rule:app-1} & \textbf{IF} \textit{product} $=$ shirt & \textbf{THEN} \textit{vendor} $=$ C \\
			\rulenumber \label{rule:app-2} & \textbf{IF} \textit{product} $=$ bag & \textbf{THEN} \textit{vendor} $\in$ \{A, B\} \\
		\end{tabular}\\
		\vspace{1em}
		\textbf{Trace:}\\
		\tikzexternalexportnextfalse
		\begin{tikzpicture}
			\node[rectangle split, rectangle split parts=2, rectangle split part align={center,center}, draw, rectangle split part fill={white!0,backOrange}] (first)
			{
				bag
				\nodepart{two}
				A
			};
			\node[rectangle split, rectangle split parts=2, rectangle split part align={center,center},
			draw, rectangle split part fill={white!0,white!0}, right=0.5 cm of first] (second)
			{
				shirt
				\nodepart{two}
				C
			};
			\node[rectangle split, rectangle split parts=2, rectangle split part align={center,center},
			draw, rectangle split part fill={white!0,backBlue}, right=0.5 cm of second] (third)
			{
				shirt
				\nodepart{two}
				B
			};
			\node[rectangle split, rectangle split parts=2, rectangle split part align={center,center},
			draw, rectangle split part fill={white!0,backPurple}, right=0.5 cm of third] (fourth)
			{
				pants
				\nodepart{two}
				C
			};
			\path[-{Latex[length=3mm,width=3mm]}] (first) edge (second);
			\path[-{Latex[length=3mm,width=3mm]}] (second) edge (third);
			\path[-{Latex[length=3mm,width=3mm]}] (third) edge (fourth);
			\node[rectangle split, rectangle split parts=2, rectangle split part align={right,right}, left=0.2 cm of first] (labels)
			{
				\textbf{product:}
				\nodepart{two}
				\textbf{vendor:}
			};
			\node[align=right, above left=0.1 cm and 0.2 cm of first] (event-label) {\textbf{event:}};
			\node[above=0.1cm of first] (event1) {\eventnumber \label{event:app-1}};
			\node[above=0.1cm of second] (event2) {\eventnumber \label{event:app-2}};
			\node[above=0.1cm of third] (event3) {\eventnumber \label{event:app-3}};
			\node[above=0.1cm of fourth] (event4) {\eventnumber \label{event:app-4}};
		\end{tikzpicture}
	\end{mdframed}
	\mycaption{Example rules and trace}{We show two simple rules (top) and a simple trace (bottom) to disprove submodularity and monotonicity for our score.}
	\label{fig:submodularity-monotonicity-example}
\end{figure}

Submodularity~\cite[p.~15]{korte:2012:combinatorial} requires our score to fulfill
\[
L(D, M_a \cap M_b) + L(D, M_a \cup M_b) \le L(D, M_a) + L(D, M_b)
\]
for all valid models $M_a, M_b$.
As a counterexample for this, consider the trace in Figure~\ref{fig:submodularity-monotonicity-example} as our event log $D$ and the models
$M_1 = \{\refrule{rule:app-1}\}$ and $M_2 = \{\refrule{rule:app-2}\}$.
Here, we get $L(D, M_1 \cap M_2) + L(D, M_1 \cup M_2) \approx 59.448$ and $L(D, M_1) + L(D, M_2) \approx 59.199$ and thus
\[
L(D, M_1 \cap M_2) + L(D, M_1 \cup M_2) > L(D, M_1) + L(D, M_2)~.
\]

\subsection{Monotonicity of our MDL score}
Monotonicity~\cite[p.~359]{korte:2012:combinatorial} requires our score to be non-increasing or non-decreasing as we add rules to our model.
To disprove this for our score, assume that our event log contains the trace in Figure~\ref{fig:submodularity-monotonicity-example} repeated 20 times.
Then, we get the scores $L(D, \emptyset) \approx 241.519$, $L(D, \{\refrule{rule:app-1}\}) \approx 279.595$ and $L(D, \{\refrule{rule:app-2}\}) \approx 239.595$.
So, extending the empty model with different rules can both increase and decrease the score, and thus it is not monotone.

\subsection{Supplementary material}
In the supplementary material\footnote{\codeurl} we make code and data publicly available.
This also includes the full Master's Thesis that the main paper is based on.
In this thesis, we provide further details on our encoding and the \ourmethod algorithm.
Further, we explain the setup of the experiments on synthetic data in more detail.